\documentclass[dvipsnames]{article} 
\usepackage{iclr2023_conference,times}

\usepackage{amsmath,amsfonts,bm}

\def\eqref#1{equation~\ref{#1}}

\def\1{\bm{1}}

\def\rvb{{\mathbf{b}}}

\def\rvx{{\mathbf{x}}}
\def\rvy{{\mathbf{y}}}
\def\rvz{{\mathbf{z}}}

\DeclareMathAlphabet{\mathsfit}{\encodingdefault}{\sfdefault}{m}{sl}
\SetMathAlphabet{\mathsfit}{bold}{\encodingdefault}{\sfdefault}{bx}{n}

\usepackage{graphicx}
\usepackage{amsmath}
\usepackage{amssymb}
\usepackage{epsfig}
\usepackage{xspace}
\usepackage{ dsfont }
\usepackage{float}
\usepackage{subcaption}
\usepackage{booktabs}
\usepackage{multicol}
\usepackage{multirow}
\usepackage[all]{nowidow}
\newcommand{\ie}[0]{\emph{i.e.\xspace~}}
\newcommand{\eg}[0]{\emph{e.g.\xspace~}}

\newcommand{\rev}[1]{\textcolor[rgb]{0,0,0}{#1}}
\newcommand{\Lcal}{\mathcal{L}}
\renewcommand{\1}{\mathds{1}}
\newcommand{\ddetr}{Def.\xspace~DETR\xspace}

\usepackage[citecolor=green,hidelinks,colorlinks=true]{hyperref}
\usepackage{url}
\usepackage[noabbrev, capitalise]{cleveref}

\title{Proposal-Contrastive Pretraining for Object Detection from Fewer Data}

\author{Quentin Bouniot$^{1,2}$, Romaric Audigier$^1$, Angélique Loesch$^1$, Amaury Habrard$^{2,3}$ \\
$^1$~Université Paris-Saclay, CEA, LIST, F-91120, Palaiseau, France \\
$^2$~Université Jean Monnet Saint-Etienne, CNRS,  Institut d Optique Graduate School,\\\phantom{$^2$}  Laboratoire Hubert Curien UMR 5516, F-42023, Saint-Etienne, France\\
$^3$~Institut Universitaire de France (IUF) \\
\texttt{firstname.lastname@\{cea.fr, univ-st-etienne.fr\}}
}

\iclrfinalcopy 
\begin{document}

\maketitle

\allowdisplaybreaks
\linepenalty=1000


\begin{abstract}
    The use of pretrained deep neural networks represents an attractive way to achieve strong results with few data available. When specialized in dense problems such as object detection, learning local rather than global information in images has proven to be more efficient. However, for unsupervised pretraining, the popular contrastive learning requires a large batch size and, therefore, a lot of resources. To address this problem, we are interested in transformer-based object detectors that have recently gained traction in the community with good performance and with the particularity of generating many diverse object proposals. 
    In this work, we present Proposal Selection Contrast (ProSeCo), a novel unsupervised overall pretraining approach that leverages this property. ProSeCo uses the large number of object proposals generated by the detector for contrastive learning, which allows the use of a smaller batch size, combined with object-level features to learn local information in the images. To improve the effectiveness of the contrastive loss, we introduce the object location information in the selection of positive examples to take into account multiple overlapping object proposals. When reusing pretrained backbone, we advocate for consistency in learning local information between the backbone and the detection head. 
    We show that our method outperforms state of the art in unsupervised pretraining for object detection on standard and novel benchmarks in learning with fewer data. 
\end{abstract}


\section{Introduction}


In recent years, we have seen a surge in research on unsupervised pretraining.
For some popular tasks such as Image Classification or Object detection, initializing with a pretrained backbone helps train big architectures more efficiently \citep{chen2020big, caron2020unsupervised, he2020momentum}. 
While gathering data is not difficult in most cases, its labeling is always time-consuming and costly.
Pretraining leverages huge amounts of unlabeled data and helps achieve better performance with fewer data and fewer iterations, when finetuning the pretrained models afterwards.

\begin{figure}[t]
    \centering
    \includegraphics[width=\linewidth]{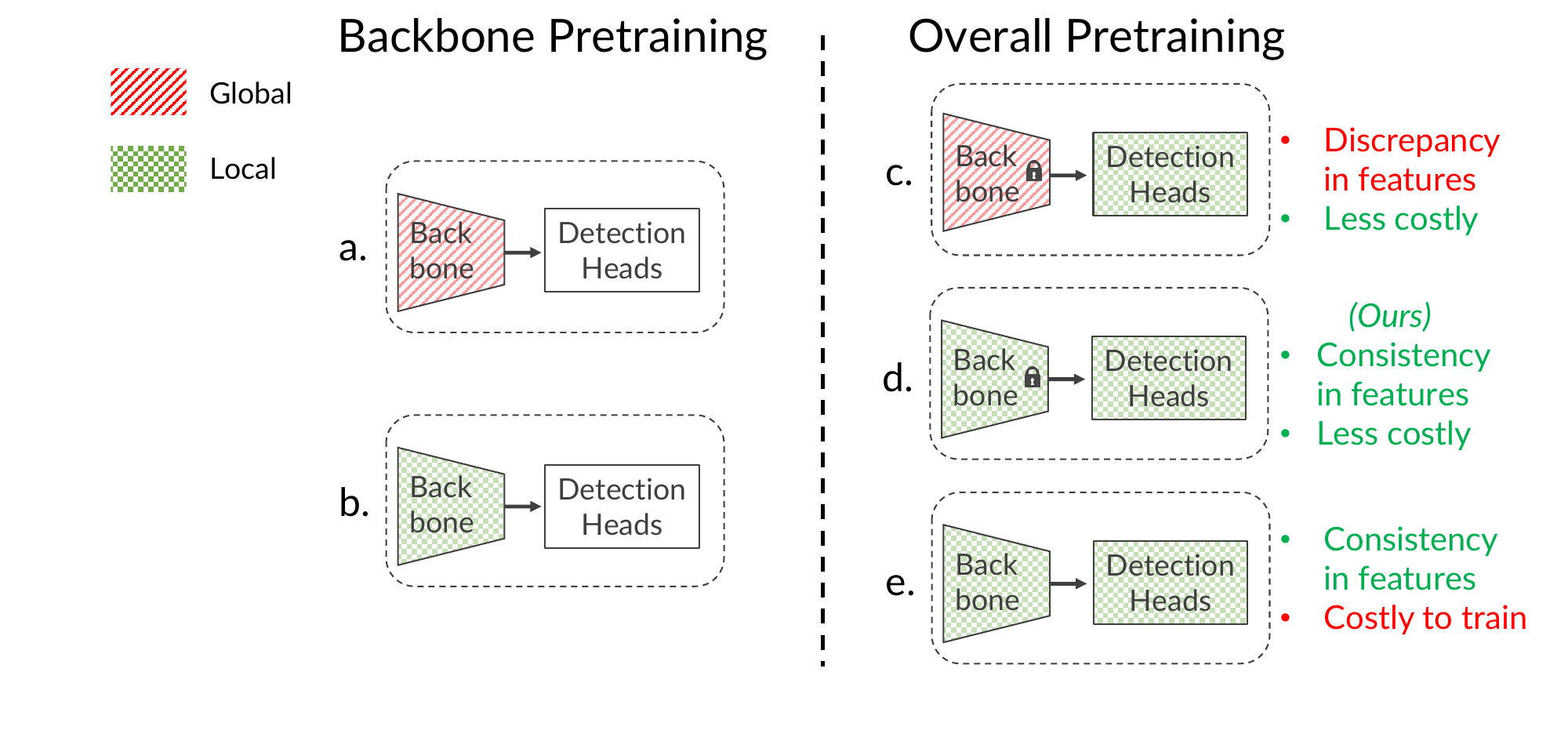}
    \caption{Illustration of the different pretraining possibilities for Object Detection. The pretraining can be either limited to the backbone (\emph{left}), or overall including the detection heads (\emph{right}). The few previous overall approaches either suffer from a discrepancy in the features between the backbone, that is trained at the image-level (\emph{global}), and the detection heads, trained to encode object-level (\emph{local}) information (\emph{c.}), or from the cost of training lots of parameters with a large batch size (\emph{e.}).}
    \label{fig:od_pt}
\end{figure}

The design of pretraining tasks for dense problems such as Object Detection has to take into account the fine-grained information in the image. Furthermore, complex object detectors contain different specific parts that can be either pretrained \emph{independently} \citep{xiao2021region, xie2021propagate, wang2021dense, henaff2021efficient, dai2021up, bar2022detreg} or \emph{jointly} \citep{wei2021aligning}. The different pretraining possibilities for Object Detection in the literature are illustrated in \cref{fig:od_pt}. A pretraining of the backbone tailored to dense tasks has been the subject of many recent efforts \citep{xiao2021region, xie2021propagate, wang2021dense, henaff2021efficient} (\emph{Backbone Pretraining}), but few have been interested in incorporating the detection-specific components of the architectures during pretraining \citep{dai2021up, bar2022detreg, wei2021aligning} (\emph{Overall Pretraining}). Among them, SoCo~\citep{wei2021aligning} focuses on convolutional architectures and pretrains the whole detector, \ie the backbone along with the detection heads (approach \emph{e.} in \cref{fig:od_pt}), whereas UP-DETR~\citep{dai2021up} and DETReg~\citep{bar2022detreg} pretrain only the transformers~\citep{vaswani2017attention} in transformer-based object detectors~\citep{carion2020end, zhu2020deformable} and keep the backbone fixed (approach \emph{c.} in \cref{fig:od_pt}).
Due to the numerous parameters that must be learned and the huge number of iterations needed because of random initialization, pretraining the entire detection model is expensive (\cref{fig:od_pt}, \emph{e.}). 
On the other hand, pretraining only the detection-specific parts with a fixed backbone leads to fewer parameters and allows leveraging strong pretrained backbones already available. However, fully relying on aligning embeddings given by the fixed backbone during pretraining and those given by the detection head, as done in DETReg or UP-DETR, introduces a discrepancy in the information contained in the features (\cref{fig:od_pt}, \emph{c.}). Indeed, while the pretrained backbone has been trained to learn image-level features, the object detector must understand object-level information in the image. \rev{Aligning inconsistent features hinders the pretraining quality.}

In this work, we propose \emph{Proposal Selection Contrast} (ProSeCo), an unsupervised pretraining method using transformer-based detectors with a fixed pretrained backbone. 
ProSeCo makes use of two models. The first one aims to alleviate the discrepancy in the features by \rev{maintaining a copy of} the whole detection model. This model is referred to as a \emph{teacher} in charge of the \emph{object proposals} embeddings, and is updated through an Exponential Moving Average (EMA) of another \emph{student} network making the object predictions and using a similar architecture. This latter network is trained by a contrastive learning approach leveraging the high number of object proposals that can be obtained from the detectors. This methodology, in addition to the absence of batch normalization in the architectures, reduces the need for a large batch size.
We further adapt the contrastive loss commonly used in pretraining to take into account the locations of the object proposals in the image, which is crucial in object detection. In addition, the localization task is independently learned through region proposals generated by Selective Search \citep{uijlings2013selective}.
Our contributions are summarized as:
\begin{itemize}
\item We propose \emph{Proposal Selection Contrast (ProSeCo)}, a contrastive learning method tailored for pretraining transformer-based object detectors. 
\item We introduce the information of the localization of object proposals for the selection of positive examples in the contrastive loss to improve its efficiency for pretraining.
\item We show that our proposed ProSeCo outperforms previous pretraining methods for transformer-based object detectors on standard benchmarks as well as novel benchmarks.
\end{itemize}

\section{Related Work}\label{sec:related_work}


\paragraph{Supervised Object Detection with transformer-based architectures}

Object Detection is an important and extensively researched problem in computer vision~\citep{girshick2014rich,girshick2015fast,ren2015faster, redmon2016you, liu2016ssd, lin2017feature, tian2019fcos}. It essentially combines object localization and classification tasks. 
Recently, a novel detector based on an encoder-decoder architecture using transformers~\citep{vaswani2017attention} has been proposed in~\citet{carion2020end}. 
The training complexity of this architecture was subsequently improved in Deformable DETR (\ddetr)~\citep{zhu2020deformable}, by changing the attention operations into deformable attention, resulting in a faster convergence speed. Several other follow-up works \citep{dai2021dynamic, meng2021conditional, wang2021anchor, liu2022dabdetr, yang2022querydet, li2022dn} have also focused on increasing the training efficiency of DETR. 
Transformer-based architectures now represent a strong alternative to traditional convolutional object detectors, reaching better performance for a similar training cost. Furthermore, recent work have shown strong results of transformer-based detectors in a data-scarce setting \citep{bar2022detreg,Bouniot_2023_WACV}, compared to convolutional architectures \citep{liu2020unbiased,Liu_2022_CVPR}, \rev{which we also observe and discuss in \cref{ax:perf_comp_fsl}.}

\paragraph{Self-supervised and unsupervised pretraining backbone architectures}

Recent advances in self-supervised pretraining~\citep{grill2020bootstrap, caron2020unsupervised, chen2020big, chen2021empirical, zheng2021ressl, denize2023similarity} have achieved strong results for obtaining general representations that transfer well to image classification~\citep{he2020momentum}. Early works proposed pretext tasks \citep{alexey2015discriminative, noroozi2016unsupervised, komodakis2018unsupervised}, which are now outperformed by \emph{Contrastive Learning}~\citep{oord2018representation, wu2018unsupervised, he2020momentum, chen2020simple, misra2020self, grill2020bootstrap, caron2020unsupervised, chen2020big, chen2021empirical, denize2023similarity}. This paradigm relies on instance discrimination using a pair of positive views from the same input contrasted with all other instances in the batch, called negatives. However, the InfoNCE objective function~\citep{oord2018representation} widely used for contrastive learning and its recent improved version, SCE~\citep{denize2023similarity}, both require a large amount of negative instances to be effective~\citep{wang2020understanding}.
The improvements observed using a general self-supervised pretraining are less significant for more complex, dense downstream tasks \citep{He_2019_ICCV,reed2022self}. To address this issue, recent approaches have started investigating pretraining approaches tailored for these tasks by imposing local consistency, either at the \emph{pixel} or \emph{region-level}: they respectively propose to match in the representation space the features corresponding to the same location in the input space \citep{o2020unsupervised, xie2021propagate, wang2021dense}, or apply local consistency between features from regions in the image \citep{roh2021spatially, yang2021instance, xiao2021region}.


\paragraph{Unsupervised Pretraining for the overall detection model}

Few approaches in the literature have tackled the problem of pretraining the detection-specific parts of the architecture, along with the backbone \citep{wei2021aligning}, or independently \citep{dai2021up, bar2022detreg}. SoCo \citep{wei2021aligning} proposes a pretraining strategy for convolutional detectors inspired by BYOL \citep{grill2020bootstrap}. Object locations are generated using Selective Search \citep{uijlings2013selective}, and then object-level features are extracted and contrasted with each other using a \emph{teacher-student} strategy. The small amount of object proposals and object features generated requires using a large batch size for the contrastive loss to be effective. Pretraining the backbone along with the detection modules this way makes the method difficult and costly to train due to the high amount of parameters to learn.
For transformer-based architectures, UP-DETR \citep{dai2021up} and DETReg \citep{bar2022detreg} use a fixed pretrained backbone to extract features respectively from random patch, or from object locations given by Selective Search \citep{uijlings2013selective}, then pretrain the detector by localizing and reconstructing the features of the patch extracted from the input images. However, since the features to reconstruct are obtained by a backbone which was trained to encode image-level information, there is a discrepancy in the information between the features to match.

Our proposed ProSeCo is designed specifically for transformer-based detectors, and use a fixed backbone pretrained for \emph{local information}. In this work, we leverage the high amount of object proposals generated by transformer-based detectors as instances for contrastive learning. Target object-level features and localizations are provided by a \emph{teacher} detection model updated through EMA, inspired by recent advances in self-supervised and semi-supervised learning \citep{liu2020unbiased, grill2020bootstrap, denize2023similarity, wei2021aligning}. The \emph{student} detection model is pretrained by computing the contrastive loss between \emph{object proposals} given by the student and teacher detectors. The large number of proposals generated by transformer-based detectors alleviates the need for a large batch size for the contrastive loss. The contrastive loss function used is further improved by introducing the location of objects for selecting positive proposals. 



\section{Overview of the approach}\label{sec:method}

\begin{figure}[t]
    \centering
    \includegraphics[width=\linewidth]{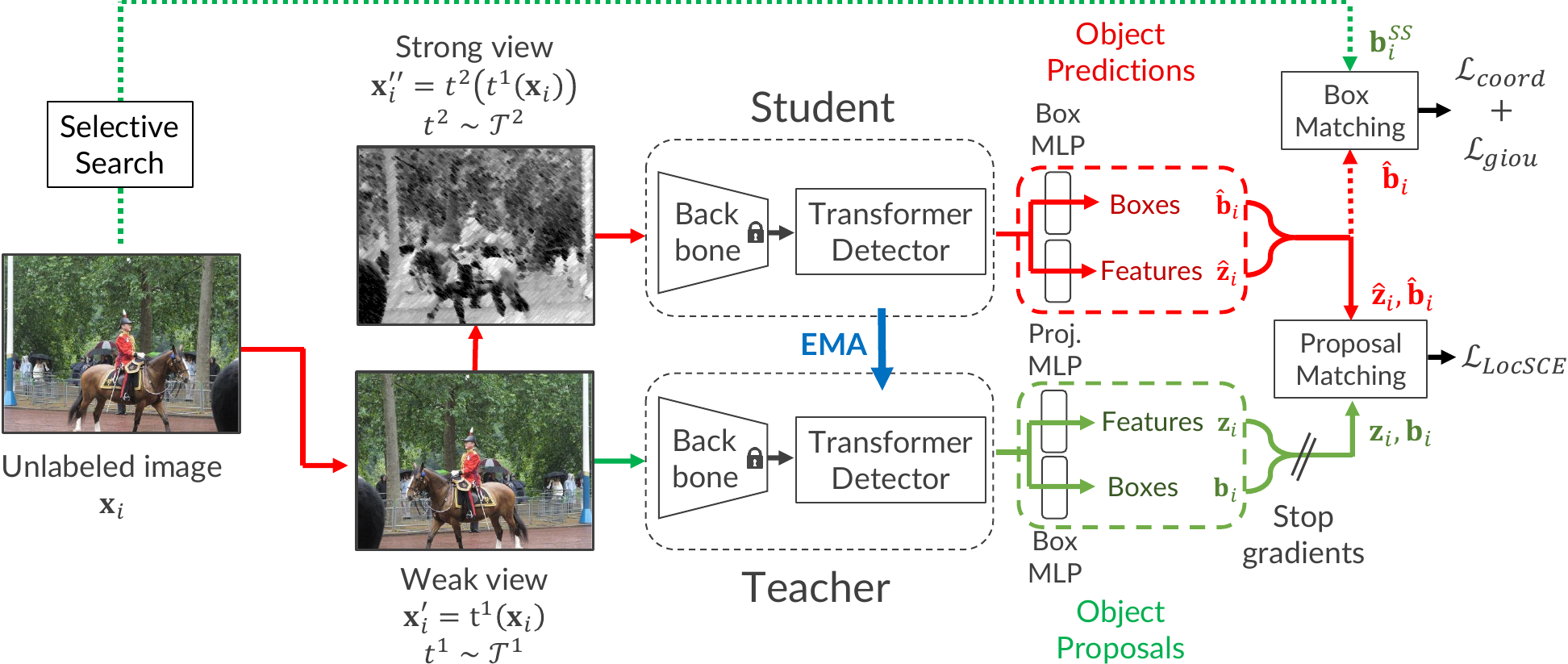}
    \caption{Overview of our proposed \emph{ProSeCo} for unsupervised pretraining. The method follows a \emph{student-teacher architecture}, with the teacher updated through an Exponential Moving Average (EMA) of the student. For each input image, $K$ random boxes are computed using the \emph{Selective Search} algorithm, and two different views are generated through an asymmetric set of \emph{weak augmentations} $\mathcal{T}^1$ and \emph{strong augmentations} $\mathcal{T}^2$. Then, \emph{object predictions} are obtained from the student model for the strongly augmented view, and \emph{object proposals} from the teacher model with the weakly augmented view. Finally, the boxes predicted by the student are matched with the boxes sampled from Selective Search to compute the localization losses $\Lcal_\text{coord}$ and $\Lcal_\text{giou}$, and the full predictions are matched with the object proposals to compute \emph{our novel contrastive loss} $\Lcal_\text{LocSCE}$.}
    \label{fig:ProSeCo}
\end{figure}

We present in this section our proposed unsupervised pretraining approach, illustrated in \cref{fig:ProSeCo}, beginning with the data processing pipeline. Then, we detail the contrastive loss used with the localization-aware positive object proposal selection. The transformer-based detectors are built on a general architecture that consists of a backbone encoder (\eg a ResNet-50), followed by several transformers encoder and decoder layers, and finally two prediction heads for the bounding boxes coordinates and class logits \citep{carion2020end, zhu2020deformable}.

\subsection{Data processing pipeline}\label{sec:data}

Throughout the section, we assume to have sampled a batch of unlabeled images $\rvx = \{\rvx_i\}_{i=1}^{N_b}$, with $\rvx_i$ the $i$\textsuperscript{th} image and $N_b$ the batch size.

For each input image $\rvx_i$, we compute two different views with two asymmetric distributions of augmentations $\mathcal{T}^1$ and $\mathcal{T}^2$: a \emph{weakly augmented view} ${\rvx_i}^\prime = t^1(\rvx_i)$, with $t^1 \sim \mathcal{T}^1$, and a \emph{strongly augmented view} obtained from the weakly augmented one ${\rvx_i}'' = t^2(t^1(\rvx_i))$, with $t^2 \sim \mathcal{T}^2$. 

To guide the model into discovering localization of objects in unlabeled images and prevent collapse, we use bounding boxes obtained from the \emph{Selective Search} algorithm \citep{uijlings2013selective}, similarly to previous work \citep{wei2021aligning, bar2022detreg}. Since Selective Search is deterministic and the generated boxes are not ordered, we compute the boxes for all images offline and, at training time, \emph{randomly} sample $K$ boxes $\rvb^{SS}_i = \{ \rvb^{SS}_{(i,j)} \in \mathbb{R}^4\}_{j=1}^K$ for each image in the batch. 
Then, the two views and the corresponding boxes sampled are used as input for our method.

\subsection{Pretraining method}\label{sec:proseco}

As shown in \cref{fig:ProSeCo}, our approach is composed of a \emph{student-teacher architecture}~\citep{grill2020bootstrap, denize2023similarity, wei2021aligning}. With ProSeCo, we extend the student-teacher pretraining for transformer-based detectors to tackle the discrepancy in information-level when aligning features, and introduce a \emph{dual unsupervised bipartite matching} presented below.

First, the backbone and the detection heads are respectively initialized from pretrained weights and randomly for both the student and teacher models. The teacher model is updated through an Exponential Moving Average (EMA) of the student's weights at every training iteration. For both models, the classification heads in the detectors are replaced by an MLP, called \emph{projector}, for obtaining latent representations of the objects. 

From the weakly augmented view ${\rvx_i}^\prime$, the teacher model provides \emph{object proposals} $\rvy_i = \{ \rvy_{(i,j)} \}_{j=1}^N = \{(\rvz_{(i,j)}, \rvb_{(i,j)} \}_{j=1}^N$, with $\rvz_{(i,j)}$ the latent embedding and $\rvb_{(i,j)}$ the coordinates of the $j$\textsuperscript{th} object found. The student model infers predictions $\hat{\rvy}_i = \{ \hat{\rvy}_{(i,j)} \}_{j=1}^N = \{(\hat{\rvz}_{(i,j)}, \hat{\rvb}_{(i,j)} \}_{j=1}^N$ from the corresponding strongly augmented view ${\rvx_i}''$.

Then, we apply an \emph{unsupervised} Hungarian algorithm for \emph{proposal matching} to find from all the permutations of $N$ elements $\mathfrak{S}_N$, the optimal bipartite matching $\hat{\sigma}_i^\text{prop}$ between the predictions $\hat{\rvy}_i$ of the student and the object proposals $\rvy_i$ of the teacher: 

\begin{equation}
\hat{\sigma}_i^\text{prop} = \arg \min_{\sigma \in \mathfrak{S}_N} \sum_{j=1}^N \Lcal_{\text{prop\_match}}(\rvy_{(i,j)}, \hat{\rvy}_{(i,\sigma(j))}).
\end{equation}

Therefore, for each image $\rvx_i$, the $j$\textsuperscript{th} proposal $\rvy_{(i,j)}$ found by the teacher is associated to the $\hat{\sigma}_i^\text{prop}(j)$\textsuperscript{th} prediction of the student $\hat{\rvy}_{(i, \hat{\sigma}_i^\text{prop}(j))}$. Our matching cost $\Lcal_{\text{prop\_match}}$ for the Hungarian algorithm takes into account both features and bounding box predictions through a linear combination of features similarity $\Lcal_\text{sim}(\rvz_{(i,j)}, \hat{\rvz}_{(i,\sigma(j))}) = \frac{ \langle \rvz_{(i,j)}, \hat{\rvz}_{(i,\sigma(j))} \rangle}{\| \rvz_{(i,j)} \|_2 \cdot \| \hat{\rvz}_{(i,\sigma(j))} \|_2} $, the $\ell_1$ loss of the box coordinates $\Lcal_{\text{coord}} = \| \rvb_{(i,j)} - \hat{\rvb}_{(i,\hat{\sigma}_i(j))} \|_1$, and the generalized IoU loss $\Lcal_{\text{giou}}$ from \citet{rezatofighi2019generalized}:

\begin{align}
\begin{split}
\Lcal_\text{prop\_match}(\rvy_{(i,j)}, \hat{\rvy}_{(i,\sigma(j))}) &= \Bigl[ \lambda_\text{sim} \Lcal_\text{sim}\left(\rvz_{(i,j)}, \hat{\rvz}_{(i,\sigma(j))}\right) \\
& + \lambda_{\text{coord}} \Lcal_\text{coord}\left(\rvb_{(i,j)}, \hat{\rvb}_{(i,\sigma(j))}\right) + \lambda_\text{giou} \Lcal_\text{giou}\left(\rvb_{(i,j)}, \hat{\rvb}_{(i,\sigma(j))}\right)\Bigr].
\end{split}
\end{align}

Similarly, we also use an \emph{unsupervised} Hungarian algorithm for \emph{box matching}, to find the optimal bipartite matching $\hat{\sigma}_i^\text{box} \in \mathfrak{S}_N$ between the predicted boxes $\hat{\rvb}_i$ of the student and the sampled boxes $\rvb^{SS}_i$ from Selective Search, using the matching cost $\Lcal_{\text{box\_match}}$:

\begin{align}
\hat{\sigma}_i^\text{box} &= \arg \min_{\sigma \in \mathfrak{S}_N} \sum_{j=1}^N \Lcal_{\text{box\_match}}(\rvy_{(i,j)}, \hat{\rvy}_{(i,\sigma(j))}), \\
\Lcal_\text{box\_match}(\rvb^{SS}_{(i,j)}, \hat{\rvb}_{(i,\sigma(j))}) &= \lambda_{\text{coord}} \Lcal_\text{coord}\left(\rvb^{SS}_{(i,j)}, \hat{\rvb}_{(i,\sigma(j))}\right) + \lambda_\text{giou} \Lcal_\text{giou}\left(\rvb^{SS}_{(i,j)}, \hat{\rvb}_{(i,\sigma(j))}\right).
\end{align}

Finally, the global unsupervised loss $\Lcal_u$ used for training is a combination of a loss function between the object latent embeddings of the teacher and student models, and between the object localization of the student predictions and Selective Search boxes. 
More formally, it is computed as:
\begin{align}
\begin{split}
    \Lcal_u(\rvx) &= \lambda_\text{contrast} \Lcal_\text{LocSCE}\left(\rvy, \hat{\rvy},\hat{\sigma}^\text{prop}\right) + \\
    & \frac{1}{N_b K} \sum_{i=1}^{N_b} \Bigl[ \sum_{j=1}^K \lambda_{\text{coord}} \Lcal_{\text{coord}}\left(\rvb^{SS}_{(i,j)}, \hat{\rvb}_{(i,\hat{\sigma}^\text{box}_i(j))}\right) + \sum_{j=1}^K \lambda_\text{giou} \Lcal_{\text{giou}}\left(\rvb^{SS}_{(i,j)}, \hat{\rvb}_{(i,\hat{\sigma}^\text{box}_i(j))}\right) \Bigr]. 
\end{split}
\end{align}

In the above equations, we define $\lambda_\text{sim}, \lambda_{\text{coord}}, \lambda_\text{giou}, \lambda_\text{contrast} \in \mathbb{R}$ as the coefficients for the different losses. For the consistency in the latent embeddings of the objects, we introduce the object locations information in our contrastive loss $\Lcal_\text{LocSCE}$. This loss is used to contrast the predictions $\hat{\rvy} = \{ \hat{\rvy}_i \}_{i=1}^{N_b}$ of the student with object proposals $\rvy = \{ \rvy_i \}_{i=1}^{N_b}$ found by the teacher, matched according to the proposal matching $\hat{\sigma}^\text{prop} = \{ \hat{\sigma}_i^\text{prop} \}_{i=1}^{N_b}$ over the batch. We detail the computations behind this loss in the next section. 

\subsection{Localization-aware contrastive loss}\label{sec:loc_sce}

\begin{figure}[t]
    \centering
    \includegraphics[width=0.75\linewidth]{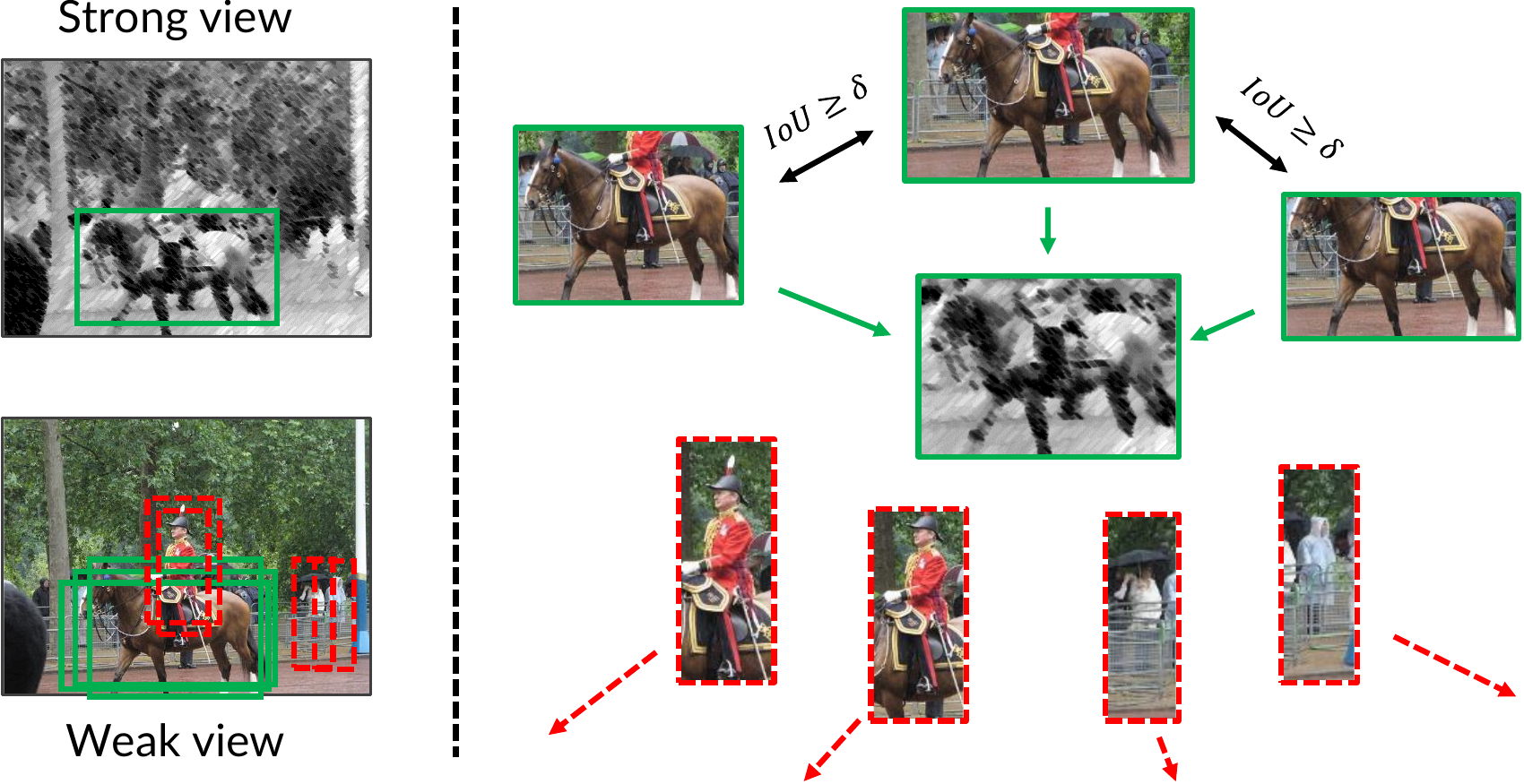}
    \caption{Illustration of the effect of the \emph{localized contrastive loss} used. Predictions of the student and teacher models are contrasted with each other to leverage the large number of object proposals obtained from transformer-based detectors. To introduce the \emph{object locations} information, overlapping proposals (\emph{in green}) in each \rev{weak view}, according to an \emph{IoU threshold} $\delta$, are also considered as positive along with the matched proposal. \rev{Proposals that neither match nor overlap the matched proposal, are considered as negative (\emph{in red}) in the contrastive loss.}}
    \label{fig:loc_sce}
\end{figure}

Inspired by advances in self-supervised learning, we propose a contrastive objective function \citep{oord2018representation, chen2020simple, he2020momentum} between object proposals that also maintains relations \citep{zheng2021ressl, denize2023similarity} among these proposals.
This objective function extends the latest SCE \citep{denize2023similarity} for \emph{instance discrimination between object proposals} and is illustrated in \cref{fig:loc_sce}. We compute the contrastive loss between all the object latent embeddings from a batch of image. The \emph{positive pair} of objects in an image $\rvx_i$ is given by the proposal matching $\sigma_i^\text{prop}$.
First, we define the distributions of similarity between objects embeddings :

\begin{align}
    p_{(in,jm)}^\prime &= \frac{ \1_{i \neq n} \1_{j \neq m} \exp(\rvz_{(i,j)} \cdot \rvz_{(n,m)} / \tau_t) }{ \sum_{k=1}^{N_b} \sum_{l=1}^{N} \1_{i \neq k} \1_{j \neq l} \exp(\rvz_{(i,j)} \cdot \rvz_{(k,l)} / \tau_t) }, \\
    p_{(in,jm)}^{\prime\prime} &= \frac{ \exp( \rvz_{(i,j)} \cdot \hat{\rvz}_{(n,m)} / \tau )}{ \sum_{k=1}^{N_b} \sum_{l=1}^{N} \exp( \rvz_{(i,j)} \cdot \hat{\rvz}_{(k,l)} / \tau )}.
\end{align}

The distribution $p_{(in,jm)}^\prime$ represents the \emph{relations} between weakly augmented object embeddings scaled by the temperature $\tau_t$, and $p_{(in,jm)}^{\prime\prime}$ the similarity between the strongly augmented embeddings and the weakly augmented ones, scaled by $\tau$.
Then, the \emph{target similarity distribution} for the objective function is a weighted combination of one-hot label and the teacher's embeddings relations:

\begin{equation}
    w_{(in,jm)} = \lambda_\text{SCE} \cdot \1_{i=n} \1_{j=m} + (1-\lambda_\text{SCE}) \cdot p_{(in,jm)}^\prime.
\end{equation}


To introduce the localization information in our objective, we compute the pairwise \emph{Intersection over Union} (IoU) between object proposals of the same image and consider overlapping objects as other positives when computing the target similarity distribution:
\begin{equation}
    w_{(in,jm)}^\text{Loc} = \lambda_\text{SCE} \cdot \1_{i=n} \1_{IoU_i(j,m) \geq \delta} + (1-\lambda_\text{SCE}) \cdot p_{(in,jm)}^\prime,
\end{equation}

\noindent where $IoU_i(j,m)$ corresponds to the IoU between teacher proposals $\rvy_{(i,j)}$ and $\rvy_{(i,m)}$ found in the same image $\rvx_i$, and $\delta$ is the \emph{IoU threshold} to consider the proposal as a positive example. 
Finally, we use this tailored target similarity distribution in our \emph{Localized SCE} (LocSCE) loss:

\begin{equation}
    \Lcal_\text{LocSCE}(\rvy, \hat{\rvy}, \hat{\sigma}^\text{prop}) = - \frac{1}{N_b N} \sum_{i=1}^{N_b} \sum_{n=1}^{N_b} \sum_{j=1}^N \sum_{m=1}^N w^\text{Loc}_{(in,jm)} \log ( p_{(in,j\hat{\sigma}_n^\text{prop}(m))}^{\prime\prime}).
\end{equation} 

Note that we do not match the proposals according to $\hat{\sigma}^\text{prop}$ in the target similarity, as we compare proposals obtained by the teacher model. We also require the full proposal as input of the loss to compute the pairwise IoU using the box coordinates. Furthermore, we recover the original formulation of SCE when $\delta =1$.
This formulation leads to an effective batch size of $N_b \cdot N$.

This localization-aware contrastive loss function aims to pull together the objects embeddings that overlap subsequently with each others, as they should correspond to the same object in the image. 

\section{Experiments}\label{sec:exps}

In this section, we present a comparative study of the results of our proposed method on standard and novel benchmarks for learning with fewer data, as well as an ablative study on the most relevant parts. First, we introduce the datasets, evaluation and training settings.

\subsection{Implementation details}\label{sec:exp_details}

\textbf{Datasets and evaluation}\phantom{.}
We use \emph{ImageNet ILSVRC 2012 (IN)} \citep{russakovsky2015imagenet} for pretraining, \emph{MS-COCO (COCO)} \citep{lin2014microsoft} and \emph{Pascal VOC 2007 and 2012} \citep{everingham2010pascal} for finetuning.
To evaluate the performance in learning with fewer data, following previous work \citep{wei2021aligning, bar2022detreg}, we consider the \emph{Mini-COCO} benchmarks, where we randomly sample 1\%, 5\% or 10\% of the training data. Similarly, we also introduce the novel \emph{Mini-VOC} benchmark, in which we randomly sample 5\% or 10\% of the training data. 
We also use the Few-Shot Object Detection (FSOD) dataset \citep{fan2020few} in the novel \emph{FSOD-test} \rev{and \emph{FSOD-train}} benchmarks. We separate the \emph{FSOD test set} with 80\% of the data randomly sampled for training and the remaining 20\% data for testing, by taking care of having at least 1 image for each class in both subsets, \rev{and do the same for the \emph{FSOD train set}}. In all benchmarks, the image ids selected for training and testing will be made available for reproducibility. More details in \cref{ax:protocols}.

\textbf{Pretraining}\phantom{.}
We initialize the backbone with the publicly available pretrained SCRL \citep{roh2021spatially} checkpoint and pretrain ProSeCo for 10 epochs on IN. The hyperparameters are set as follows: the EMA keep rate parameter to 0.999, the IoU threshold $\delta = 0.5$, a batch size of $N_b = 64$ images over 8 A100 GPUs, and the coefficients in the different losses $\lambda_\text{sim} = \lambda_\text{contrast} = 2$ which is the same value used for the coefficient governing the class cross-entropy in the supervised loss. The projector is defined as a 2-layer MLP with a hidden layer of 4096 and a last layer of 256, without batch normalization. Following SCE \citep{denize2023similarity}, we set the temperatures $\tau = 0.1, \tau_t = 0.07$ and the coefficient $\lambda_\text{SCE} = 0.5$. We sample $K=30$ \emph{random} boxes from the outputs of Selective Search for each image at every iteration. Other training and architecture hyperparameters are defined as in \ddetr \citep{zhu2020deformable} with, specifically, the coefficients $\lambda_\text{coord} = 5$ and $\lambda_\text{giou} = 2$, the number of object proposals (queries) $N=300$, and the learning rate is set to $lr = 2\cdot 10^{-4}$. For weak augmentations $\mathcal{T}^1$, we use a random combination of flip, resize and crop, and for strong augmentations $\mathcal{T}^2$, we use a random combination of color jittering, grayscale and Gaussian blur. In $\mathcal{T}^1$, we resize images with the same range of scales as the supervised training protocol on COCO (\emph{Large-scale}).
The exact parameters for the augmentations are detailed in \cref{ax:augm}, \rev{and a discussion about the pretraining cost can be found in \cref{ax:cost}.}

\textbf{Finetuning protocols}\phantom{.}
For finetuning the pretrained models, we follow the standard supervised learning hyperparameters of \ddetr \citep{zhu2020deformable}. In all experiments, we train the models with a batch size of 32 images over 8 A100 GPUs until the validation performance stops increasing, \ie for Mini-COCO, up to 2000 epochs for 1\%, 500 epochs for 5\%, 400 epochs for 10\%, for Mini-VOC, up to 2000 epochs for both 5\% and 10\%, up to 100 epochs for both FSOD-test and PASCAL VOC, \rev{and up to 50 epochs for FSOD-train}. We always decay the learning rates by a factor of 0.1 after about 80\% of training.
\rev{Experiments with more annotated data are discussed in \cref{ax:ft_full_coco}.}
To compare our method to DETReg \citep{bar2022detreg} on our novel benchmarks, we use their publicly available checkpoints from github.

\subsection{Finetuning and transfer learning}

We evaluate the transfer learning ability of our pretrained model on several datasets.
\cref{tab:mini_coco,tab:voc_fsod} present the results obtained compared to previous methods in the literature when learning from fewer labeled data. We can see that our method outperforms state-of-the-art results in unsupervised pretraining on all benchmarks and datasets, and obtain even more strong results when training data is scarce. The improvement is even more significant as the overall performance with few training data is low.
When using 5\% of the COCO training data (\ie Mini-COCO 5\% in \cref{tab:mini_coco}), corresponding to about 5.9k images, ProSeCo achieves 28.8 mAP, which represents an improvement of +5.2 \emph{percentage point} (p.p.) over the supervised pretraining baseline and +2 p.p. over both state-of-the-art overall pretraining methods. \rev{Results with all the evaluation metrics are presented in \cref{ax:full_eval}.}

\begin{table}[ht]
        \begin{center}
        \caption{Performance (mAP in \%) of our proposed pretraining approach after finetuning using different percentage of training data (with the corresponding number of images reported). We show that our ProSeCo outperforms previous pretraining methods in all benchmarks.}
        \end{center}
    \subfloat[][\label{tab:mini_coco}]{
    \resizebox{0.98\linewidth}{!}{%
    \begin{tabular}{@{}lllccc@{}}
    \toprule
    \multirow{2}{*}{Method} & \multirow{2}{*}{Detector} & Pretrain. & \multicolumn{3}{c}{Mini-COCO} \\
    \cmidrule(lr){4-6}
    & & Dataset & 1\% (1.2k) & 5\% (5.9k) & 10\% (11.8k)\\
    \midrule
        Supervised & \ddetr & IN & 13.0 & 23.6 & 28.6 \\
        SwAV \citep{caron2020unsupervised} & \ddetr & IN & 13.3 & 24.5 & 29.5 \\ 
        SCRL \citep{roh2021spatially} & \ddetr & IN & 16.4 & 26.2 & 30.6 \\ 
        DETReg \citep{bar2022detreg} & \ddetr & COCO & 15.8 & 26.7 & 30.7 \\
        DETReg \citep{bar2022detreg} & \ddetr & IN & 15.9 & 26.1 & 30.9 \\
    \midrule
        Supervised \citep{wei2021aligning} & Mask R-CNN & IN & -- & 19.4 & 24.7 \\
        SoCo$^*$ \citep{wei2021aligning} & Mask R-CNN & IN & -- & 26.8 & 31.1 \\ 
    \midrule
        \textit{ProSeCo (Ours)} & \ddetr & IN & \textbf{18.0} & \textbf{28.8} & \textbf{32.8} \\
    \bottomrule
    \end{tabular}%
    }
    }
    
    \centering
    \subfloat[][\label{tab:voc_fsod}]{
    \resizebox{0.98\linewidth}{!}{%
    \begin{tabular}{@{}lccccc@{}}
    \toprule
    \multirow{2}{*}{Method} & FSOD-test & \rev{FSOD-train} & PASCAL VOC & \multicolumn{2}{c}{Mini-VOC} \\
    \cmidrule(lr){5-6}
     & 100\% (11k) & \rev{100\% (42k)} & 100\% (16k) & 5\% (0.8k) & 10\% (1.6k) \\
    \midrule
        Supervised & 39.3 & \rev{42.6} & 59.5 & 33.9 & 40.8 \\
        DETReg \citep{bar2022detreg} & 43.2 & \rev{43.3} & 63.5 & 43.1 & 48.2 \\
        \textit{ProSeCo (Ours)} & \textbf{46.6} & \rev{\textbf{47.2}} & \textbf{65.1} & \textbf{46.1} & \textbf{51.3} \\
    \bottomrule
    \end{tabular}%
    }
    }
\end{table}

\subsection{Ablation Studies}

In the following, we provide several ablation studies for our proposed approach. All experiments and results are compared on the Mini-COCO 5\% benchmark \rev{with the pretrained SwAV backbone unless explicitly stated}. \rev{Additional ablation on the number of queries can be found in \cref{ax:queries}.} 

\textbf{Pretraining dataset and backbone}\phantom{.} In \cref{tab:abl_bb}, we show the effect of changing the pretraining dataset or the fixed backbone used. We can see that using a backbone more adapted to dense tasks that learned local information (\eg SCRL) helps the model by having consistent features (+1 p.p.), compared to a backbone pretrained for global features (\eg SwAV). Furthermore, even with a less adapted backbone, our ProSeCo initialized with the SwAV backbone outperforms DETReg (+1.7 p.p.). Pretraining with DETReg improves when using a more adapted backbone, but ProSeCo still reaches better performance. To compare with IN, we also pretrain ProSeCo on COCO for 120 epochs. We obtain better results when pretraining the model on IN than using COCO thanks to the large number of different images in IN (about 10 times the number of images of COCO, leading to +0.4 p.p.), which is consistent with previous findings \citep{wei2021aligning}. 

\textbf{Localization information in contrastive loss}\phantom{.} In \cref{tab:abl_iou}, we show the effect of the localization information in the contrastive loss SCE. We can see that when introducing multiple positive examples for each image based on the IoU threshold $\delta$ (\ie $\forall \delta < 1$), we achieve better results than with the original SCE loss (\ie $\delta = 1$). Notably, the best results are achieved with $\delta = 0.5$ (+1.7 p.p.). More experiments with the InfoNCE loss \citep{oord2018representation} can be found in \cref{ax:nce}.

\textbf{Hyperparameters}\phantom{.} \cref{tab:abl_hp} presents an ablation study on different important hyperparameters of our approach.
We experimented first with the same batch size applied in \citet{bar2022detreg} (\emph{Abl.~Batch}), but found that using a smaller batch size (\emph{Base}) leads to improved results (+0.2 p.p.). 
We evaluated different image scales as a parameter of the weak data augmentations distribution in \emph{Abl.~Scale}. \emph{Mid-scale} corresponds to a resizing of the images such that the shortest edge is between 320 and 480 pixels, as used in previous work \citep{dai2021up, bar2022detreg}, and \emph{Large-scale} to a resize between 480 and 800 pixels, used for supervised learning on COCO. Exact values for these parameters can be found in \cref{ax:augm}. We found that increasing the size of the images during pretraining is important to have more meaningful information in the boxes, and a more precise localization of the boxes (+0.7 p.p.).
Following the best results from \citet{denize2023similarity}, we evaluated the performance for $\tau_t \in \{0.05, 0.07\}$. We found that $\tau_t = 0.07$ leads to the best performance (+0.3 p.p.).
We considered several EMA keep rate parameter values following previous work \citep{he2020momentum, wei2021aligning, denize2023similarity}, and found that 0.999 achieves the best results (+0.1 p.p.).

\begin{table}[ht]
    \begin{center}
        \caption{(a) Comparison after finetuning when using different pretrained backbone and/or pretraining datasets. (b) Comparison of the effect of the localization information using different IoU threshold $\delta$. All performance (mAP in \%) are measured on Mini-COCO 5\%.}
    \end{center}
    \centering
    \begin{minipage}[c]{0.59\linewidth}
    \subfloat[][\label{tab:abl_bb}]{
    \centering
    \begin{tabular}{@{}llc@{}}
    \toprule
    Pretraining & Dataset & mAP \\
    \midrule
        ProSeCo w/ SwAV  & COCO & 27.4 \\
        ProSeCo w/ SwAV  & IN & 27.8 \\
        DETReg w/ SCRL & IN & 28.0 \\
        ProSeCo w/ SCRL  & IN & \textbf{28.8} \\
    \bottomrule
    \end{tabular}%
    }
    \end{minipage}
    \begin{minipage}[c]{0.35\linewidth}
    \subfloat[][\label{tab:abl_iou}]{
    \centering
    \begin{tabular}{@{}lcc@{}}
    \toprule
    Loss & $\delta$ & mAP \\
    \midrule
        SCE & 1.0 & 26.1 \\
        \textit{LocSCE (Ours)} & 0.2 & 27.0 \\
        \textit{LocSCE (Ours)} & 0.7 & 27.1 \\
        \textit{LocSCE (Ours)} & 0.5 & \textbf{27.8} \\
    \bottomrule
    \end{tabular}
    }
    \end{minipage}
\end{table}

\begin{table}[ht]
    \begin{center}
        \caption{Ablation studies on different hyperparameters for the proposed method. The performance (mAP in \%) are measured on Mini-COCO 5\%. \textcolor{ForestGreen}{\textbf{Green and bold columns names}} indicate a \emph{positive} effect on the performance, and \textcolor{red}{red columns} a \emph{negative} effect.}
        \label{tab:abl_hp}
    \end{center}
    \centering
    \begin{tabular}{@{}lcccccccccc@{}}
    \toprule
    \multirow{2}{*}{Ablative Variant} & \multicolumn{2}{c}{Batch size} & \multicolumn{2}{c}{Images scale} & \multicolumn{2}{c}{Temperature $\tau_t$} & \multicolumn{3}{c}{EMA} & \multirow{2}{*}{mAP} \\
    \cmidrule(lr){2-3}
    \cmidrule(lr){4-5}
    \cmidrule(lr){6-7}
    \cmidrule(lr){8-10}
    & \textcolor{red}{192} & \textcolor{ForestGreen}{\textbf{64}} & \textcolor{red}{Mid} & \textcolor{ForestGreen}{\textbf{Large}} & \textcolor{red}{0.05} & \textcolor{ForestGreen}{\textbf{0.07}} & \textcolor{red}{0.99} & \textcolor{red}{0.996} & \textcolor{ForestGreen}{\textbf{0.999}} & \\
    \midrule
    Base & & \checkmark & \checkmark & & \checkmark & & & \checkmark & & 26.7 \\
    \midrule
    Abl.~Batch & \checkmark & & \checkmark & & \checkmark & & & \checkmark & & 26.5 \\
    \midrule
    Abl.~Scale &  & \checkmark & & \checkmark & \checkmark & & & \checkmark & & 27.4 \\
    \midrule
    Abl.~Temp. & & \checkmark & \checkmark & & & \checkmark & & \checkmark & & 27 \\
    \midrule
    \multirow{2}{*}{Abl.~EMA} & & \checkmark & \checkmark & & \checkmark & & \checkmark & & & 26.3 \\
    & & \checkmark & \checkmark & & \checkmark & & & & \checkmark & 26.8 \\
    \midrule
    \textbf{Best} & & \checkmark & & \checkmark & & \checkmark & & & \checkmark & \textbf{27.8} \\
    \bottomrule
    \end{tabular}%
\end{table}

\section{Conclusion}

In this work, we aim to use large unlabeled datasets for an unsupervised pretraining of the overall detection model to improve performance when having access to fewer labeled data. In this end, we propose \emph{Proposal Selection Contrast (ProSeCo)}, a novel pretraining approach for Object Detection. Our method leverages the large number of object proposals generated by transformer-based detectors for contrastive learning, reducing the necessity of a large batch size, and introducing the localization information of the objects for the selection of positive examples to contrast. We show from various experiments on standard and novel benchmarks in learning with few training data that ProSeCo outperforms previous pretraining methods. Throughout this work, we advocate for consistency in the level of information encoded in the features when pretraining. Indeed, learning object-level features during pretraining is more important than image-level when applied to a dense downstream task such as Object Detection.
Future work could update the backbone during pretraining to further improve the consistency between the backbone and the detection heads.


\section*{Reproducibility Statement}

Throughout this paper, we made sure that the experiments and results are fully reproducible. We explicitly state the exact values of hyperparameters used in \cref{sec:exp_details}, and describe in details the datasets and evaluation protocols in \cref{ax:protocols}. All image ids randomly selected when evaluating with few training data (in Mini-COCO, Mini-VOC, FSOD-test, \rev{FSOD-train}) will be made available.

\section*{Acknowledgements}

This work was made possible by the use of the Factory-AI supercomputer, financially supported by the Ile-de-France Regional Council, and also performed using HPC resources from GENCI-IDRIS (Grant 2022-AD011013478). 

\bibliography{references}

\begin{thebibliography}{53}
\providecommand{\natexlab}[1]{#1}
\providecommand{\url}[1]{\texttt{#1}}
\expandafter\ifx\csname urlstyle\endcsname\relax
  \providecommand{\doi}[1]{doi: #1}\else
  \providecommand{\doi}{doi: \begingroup \urlstyle{rm}\Url}\fi

\bibitem[Alexey et~al.(2015)Alexey, Fischer, Tobias, Springenberg, and
  Brox]{alexey2015discriminative}
Dosovitskiy Alexey, Philipp Fischer, Jost Tobias, Martin~Riedmiller
  Springenberg, and Thomas Brox.
\newblock Discriminative unsupervised feature learning with exemplar
  convolutional neural networks.
\newblock \emph{IEEE Trans. Pattern Analysis and Machine Intelligence}, 2015.

\bibitem[Bar et~al.(2022)Bar, Wang, Kantorov, Reed, Herzig, Chechik, Rohrbach,
  Darrell, and Globerson]{bar2022detreg}
Amir Bar, Xin Wang, Vadim Kantorov, Colorado~J Reed, Roei Herzig, Gal Chechik,
  Anna Rohrbach, Trevor Darrell, and Amir Globerson.
\newblock {DETReg}: Unsupervised pretraining with region priors for object
  detection.
\newblock In \emph{Proceedings of the IEEE/CVF Conference on Computer Vision
  and Pattern Recognition}, 2022.

\bibitem[Bouniot et~al.(2023)Bouniot, Loesch, Audigier, and
  Habrard]{Bouniot_2023_WACV}
Quentin Bouniot, Ang\'elique Loesch, Romaric Audigier, and Amaury Habrard.
\newblock Towards few-annotation learning for object detection: Are
  transformer-based models more efficient?
\newblock In \emph{Proceedings of the IEEE/CVF Winter Conference on
  Applications of Computer Vision (WACV)}, pp.\  75--84, January 2023.

\bibitem[Cai et~al.(2020)Cai, Frankle, Schwab, and Morcos]{cai2020all}
Tiffany~Tianhui Cai, Jonathan Frankle, David~J Schwab, and Ari~S Morcos.
\newblock Are all negatives created equal in contrastive instance
  discrimination?
\newblock \emph{arXiv preprint arXiv:2010.06682}, 2020.

\bibitem[Carion et~al.(2020)Carion, Massa, Synnaeve, Usunier, Kirillov, and
  Zagoruyko]{carion2020end}
Nicolas Carion, Francisco Massa, Gabriel Synnaeve, Nicolas Usunier, Alexander
  Kirillov, and Sergey Zagoruyko.
\newblock End-to-end object detection with transformers.
\newblock In \emph{European conference on computer vision}. Springer, 2020.

\bibitem[Caron et~al.(2020)Caron, Misra, Mairal, Goyal, Bojanowski, and
  Joulin]{caron2020unsupervised}
Mathilde Caron, Ishan Misra, Julien Mairal, Priya Goyal, Piotr Bojanowski, and
  Armand Joulin.
\newblock Unsupervised learning of visual features by contrasting cluster
  assignments.
\newblock \emph{Advances in Neural Information Processing Systems}, 2020.

\bibitem[Chen et~al.(2020{\natexlab{a}})Chen, Kornblith, Norouzi, and
  Hinton]{chen2020simple}
Ting Chen, Simon Kornblith, Mohammad Norouzi, and Geoffrey Hinton.
\newblock A simple framework for contrastive learning of visual
  representations.
\newblock In \emph{International conference on machine learning}. PMLR,
  2020{\natexlab{a}}.

\bibitem[Chen et~al.(2020{\natexlab{b}})Chen, Kornblith, Swersky, Norouzi, and
  Hinton]{chen2020big}
Ting Chen, Simon Kornblith, Kevin Swersky, Mohammad Norouzi, and Geoffrey~E
  Hinton.
\newblock Big self-supervised models are strong semi-supervised learners.
\newblock \emph{Advances in neural information processing systems},
  2020{\natexlab{b}}.

\bibitem[Chen et~al.(2021)Chen, Xie, and He]{chen2021empirical}
Xinlei Chen, Saining Xie, and Kaiming He.
\newblock An empirical study of training self-supervised vision transformers.
\newblock In \emph{Proceedings of the IEEE/CVF International Conference on
  Computer Vision}, 2021.

\bibitem[Dai et~al.(2021{\natexlab{a}})Dai, Chen, Yang, Zhang, Yuan, and
  Zhang]{dai2021dynamic}
Xiyang Dai, Yinpeng Chen, Jianwei Yang, Pengchuan Zhang, Lu~Yuan, and Lei
  Zhang.
\newblock Dynamic {DETR}: End-to-end object detection with dynamic attention.
\newblock In \emph{Proceedings of the IEEE/CVF International Conference on
  Computer Vision}, 2021{\natexlab{a}}.

\bibitem[Dai et~al.(2021{\natexlab{b}})Dai, Cai, Lin, and Chen]{dai2021up}
Zhigang Dai, Bolun Cai, Yugeng Lin, and Junying Chen.
\newblock Up-{DETR}: Unsupervised pre-training for object detection with
  transformers.
\newblock In \emph{Proceedings of the IEEE/CVF conference on computer vision
  and pattern recognition}, 2021{\natexlab{b}}.

\bibitem[Denize et~al.(2023)Denize, Rabarisoa, Orcesi, H{\'e}rault, and
  Canu]{denize2023similarity}
Julien Denize, Jaonary Rabarisoa, Astrid Orcesi, Romain H{\'e}rault, and
  St{\'e}phane Canu.
\newblock Similarity contrastive estimation for self-supervised soft
  contrastive learning.
\newblock In \emph{Proceedings of the IEEE/CVF Winter Conference on
  Applications of Computer Vision}, pp.\  2706--2716, 2023.

\bibitem[Everingham et~al.(2010)Everingham, Van~Gool, Williams, Winn, and
  Zisserman]{everingham2010pascal}
Mark Everingham, Luc Van~Gool, Christopher~KI Williams, John Winn, and Andrew
  Zisserman.
\newblock The pascal visual object classes (voc) challenge.
\newblock \emph{International journal of computer vision}, 2010.

\bibitem[Fan et~al.(2020)Fan, Zhuo, Tang, and Tai]{fan2020few}
Qi~Fan, Wei Zhuo, Chi-Keung Tang, and Yu-Wing Tai.
\newblock Few-shot object detection with attention-rpn and multi-relation
  detector.
\newblock In \emph{Proceedings of the IEEE/CVF conference on computer vision
  and pattern recognition}, pp.\  4013--4022, 2020.

\bibitem[Girshick(2015)]{girshick2015fast}
Ross Girshick.
\newblock Fast r-cnn.
\newblock In \emph{Proceedings of the IEEE international conference on computer
  vision}, 2015.

\bibitem[Girshick et~al.(2014)Girshick, Donahue, Darrell, and
  Malik]{girshick2014rich}
Ross Girshick, Jeff Donahue, Trevor Darrell, and Jitendra Malik.
\newblock Rich feature hierarchies for accurate object detection and semantic
  segmentation.
\newblock In \emph{Proceedings of the IEEE conference on computer vision and
  pattern recognition}, 2014.

\bibitem[Grill et~al.(2020)Grill, Strub, Altch{\'e}, Tallec, Richemond,
  Buchatskaya, Doersch, Avila~Pires, Guo, Gheshlaghi~Azar,
  et~al.]{grill2020bootstrap}
Jean-Bastien Grill, Florian Strub, Florent Altch{\'e}, Corentin Tallec, Pierre
  Richemond, Elena Buchatskaya, Carl Doersch, Bernardo Avila~Pires, Zhaohan
  Guo, Mohammad Gheshlaghi~Azar, et~al.
\newblock Bootstrap your own latent-a new approach to self-supervised learning.
\newblock \emph{Advances in neural information processing systems}, 2020.

\bibitem[He et~al.(2019)He, Girshick, and Dollar]{He_2019_ICCV}
Kaiming He, Ross Girshick, and Piotr Dollar.
\newblock Rethinking imagenet pre-training.
\newblock In \emph{Proceedings of the IEEE/CVF International Conference on
  Computer Vision (ICCV)}, 2019.

\bibitem[He et~al.(2020)He, Fan, Wu, Xie, and Girshick]{he2020momentum}
Kaiming He, Haoqi Fan, Yuxin Wu, Saining Xie, and Ross Girshick.
\newblock Momentum contrast for unsupervised visual representation learning.
\newblock In \emph{Proceedings of the IEEE/CVF conference on computer vision
  and pattern recognition}, 2020.

\bibitem[H{\'e}naff et~al.(2021)H{\'e}naff, Koppula, Alayrac, Van~den Oord,
  Vinyals, and Carreira]{henaff2021efficient}
Olivier~J H{\'e}naff, Skanda Koppula, Jean-Baptiste Alayrac, Aaron Van~den
  Oord, Oriol Vinyals, and Jo{\~a}o Carreira.
\newblock Efficient visual pretraining with contrastive detection.
\newblock In \emph{Proceedings of the IEEE/CVF International Conference on
  Computer Vision}, 2021.

\bibitem[Komodakis \& Gidaris(2018)Komodakis and
  Gidaris]{komodakis2018unsupervised}
Nikos Komodakis and Spyros Gidaris.
\newblock Unsupervised representation learning by predicting image rotations.
\newblock In \emph{International Conference on Learning Representations
  (ICLR)}, 2018.

\bibitem[Li et~al.(2022)Li, Zhang, Liu, Guo, Ni, and Zhang]{li2022dn}
Feng Li, Hao Zhang, Shilong Liu, Jian Guo, Lionel~M Ni, and Lei Zhang.
\newblock Dn-{DETR}: Accelerate {DETR} training by introducing query denoising.
\newblock In \emph{Proceedings of the IEEE/CVF Conference on Computer Vision
  and Pattern Recognition}, 2022.

\bibitem[Lin et~al.(2014)Lin, Maire, Belongie, Hays, Perona, Ramanan,
  Doll{\'a}r, and Zitnick]{lin2014microsoft}
Tsung-Yi Lin, Michael Maire, Serge Belongie, James Hays, Pietro Perona, Deva
  Ramanan, Piotr Doll{\'a}r, and C~Lawrence Zitnick.
\newblock Microsoft coco: Common objects in context.
\newblock In \emph{European conference on computer vision}. Springer, 2014.

\bibitem[Lin et~al.(2017)Lin, Doll{\'a}r, Girshick, He, Hariharan, and
  Belongie]{lin2017feature}
Tsung-Yi Lin, Piotr Doll{\'a}r, Ross Girshick, Kaiming He, Bharath Hariharan,
  and Serge Belongie.
\newblock Feature pyramid networks for object detection.
\newblock In \emph{Proceedings of the IEEE conference on computer vision and
  pattern recognition}, 2017.

\bibitem[Liu et~al.(2022{\natexlab{a}})Liu, Li, Zhang, Yang, Qi, Su, Zhu, and
  Zhang]{liu2022dabdetr}
Shilong Liu, Feng Li, Hao Zhang, Xiao Yang, Xianbiao Qi, Hang Su, Jun Zhu, and
  Lei Zhang.
\newblock {DAB}-{DETR}: Dynamic anchor boxes are better queries for {DETR}.
\newblock In \emph{International Conference on Learning Representations},
  2022{\natexlab{a}}.

\bibitem[Liu et~al.(2016)Liu, Anguelov, Erhan, Szegedy, Reed, Fu, and
  Berg]{liu2016ssd}
Wei Liu, Dragomir Anguelov, Dumitru Erhan, Christian Szegedy, Scott Reed,
  Cheng-Yang Fu, and Alexander~C Berg.
\newblock Ssd: Single shot multibox detector.
\newblock In \emph{European conference on computer vision}. Springer, 2016.

\bibitem[Liu et~al.(2020)Liu, Ma, He, Kuo, Chen, Zhang, Wu, Kira, and
  Vajda]{liu2020unbiased}
Yen-Cheng Liu, Chih-Yao Ma, Zijian He, Chia-Wen Kuo, Kan Chen, Peizhao Zhang,
  Bichen Wu, Zsolt Kira, and Peter Vajda.
\newblock Unbiased teacher for semi-supervised object detection.
\newblock In \emph{International Conference on Learning Representations}, 2020.

\bibitem[Liu et~al.(2022{\natexlab{b}})Liu, Ma, and Kira]{Liu_2022_CVPR}
Yen-Cheng Liu, Chih-Yao Ma, and Zsolt Kira.
\newblock Unbiased teacher v2: Semi-supervised object detection for anchor-free
  and anchor-based detectors.
\newblock In \emph{Proceedings of the IEEE/CVF Conference on Computer Vision
  and Pattern Recognition (CVPR)}, pp.\  9819--9828, June 2022{\natexlab{b}}.

\bibitem[Meng et~al.(2021)Meng, Chen, Fan, Zeng, Li, Yuan, Sun, and
  Wang]{meng2021conditional}
Depu Meng, Xiaokang Chen, Zejia Fan, Gang Zeng, Houqiang Li, Yuhui Yuan, Lei
  Sun, and Jingdong Wang.
\newblock Conditional {DETR} for fast training convergence.
\newblock In \emph{Proceedings of the IEEE/CVF International Conference on
  Computer Vision}, 2021.

\bibitem[Misra \& Maaten(2020)Misra and Maaten]{misra2020self}
Ishan Misra and Laurens van~der Maaten.
\newblock Self-supervised learning of pretext-invariant representations.
\newblock In \emph{Proceedings of the IEEE/CVF Conference on Computer Vision
  and Pattern Recognition}, 2020.

\bibitem[Noroozi \& Favaro(2016)Noroozi and Favaro]{noroozi2016unsupervised}
Mehdi Noroozi and Paolo Favaro.
\newblock Unsupervised learning of visual representations by solving jigsaw
  puzzles.
\newblock In \emph{European conference on computer vision}. Springer, 2016.

\bibitem[O~Pinheiro et~al.(2020)O~Pinheiro, Almahairi, Benmalek, Golemo, and
  Courville]{o2020unsupervised}
Pedro~O O~Pinheiro, Amjad Almahairi, Ryan Benmalek, Florian Golemo, and Aaron~C
  Courville.
\newblock Unsupervised learning of dense visual representations.
\newblock \emph{Advances in Neural Information Processing Systems}, 2020.

\bibitem[Oord et~al.(2018)Oord, Li, and Vinyals]{oord2018representation}
Aaron van~den Oord, Yazhe Li, and Oriol Vinyals.
\newblock Representation learning with contrastive predictive coding.
\newblock \emph{arXiv preprint arXiv:1807.03748}, 2018.

\bibitem[Redmon et~al.(2016)Redmon, Divvala, Girshick, and
  Farhadi]{redmon2016you}
Joseph Redmon, Santosh Divvala, Ross Girshick, and Ali Farhadi.
\newblock You only look once: Unified, real-time object detection.
\newblock In \emph{Proceedings of the IEEE conference on computer vision and
  pattern recognition}, 2016.

\bibitem[Reed et~al.(2022)Reed, Yue, Nrusimha, Ebrahimi, Vijaykumar, Mao, Li,
  Zhang, Guillory, Metzger, et~al.]{reed2022self}
Colorado~J Reed, Xiangyu Yue, Ani Nrusimha, Sayna Ebrahimi, Vivek Vijaykumar,
  Richard Mao, Bo~Li, Shanghang Zhang, Devin Guillory, Sean Metzger, et~al.
\newblock Self-supervised pretraining improves self-supervised pretraining.
\newblock In \emph{Proceedings of the IEEE/CVF Winter Conference on
  Applications of Computer Vision}, 2022.

\bibitem[Ren et~al.(2015)Ren, He, Girshick, and Sun]{ren2015faster}
Shaoqing Ren, Kaiming He, Ross Girshick, and Jian Sun.
\newblock Faster r-cnn: Towards real-time object detection with region proposal
  networks.
\newblock \emph{Advances in neural information processing systems}, 2015.

\bibitem[Rezatofighi et~al.(2019)Rezatofighi, Tsoi, Gwak, Sadeghian, Reid, and
  Savarese]{rezatofighi2019generalized}
Hamid Rezatofighi, Nathan Tsoi, JunYoung Gwak, Amir Sadeghian, Ian Reid, and
  Silvio Savarese.
\newblock Generalized intersection over union: A metric and a loss for bounding
  box regression.
\newblock In \emph{Proceedings of the IEEE/CVF conference on computer vision
  and pattern recognition}, 2019.

\bibitem[Roh et~al.(2021)Roh, Shin, Kim, and Kim]{roh2021spatially}
Byungseok Roh, Wuhyun Shin, Ildoo Kim, and Sungwoong Kim.
\newblock Spatially consistent representation learning.
\newblock In \emph{Proceedings of the IEEE/CVF Conference on Computer Vision
  and Pattern Recognition}, 2021.

\bibitem[Russakovsky et~al.(2015)Russakovsky, Deng, Su, Krause, Satheesh, Ma,
  Huang, Karpathy, Khosla, Bernstein, et~al.]{russakovsky2015imagenet}
Olga Russakovsky, Jia Deng, Hao Su, Jonathan Krause, Sanjeev Satheesh, Sean Ma,
  Zhiheng Huang, Andrej Karpathy, Aditya Khosla, Michael Bernstein, et~al.
\newblock Imagenet large scale visual recognition challenge.
\newblock \emph{International journal of computer vision}, 2015.

\bibitem[Tian et~al.(2019)Tian, Shen, Chen, and He]{tian2019fcos}
Zhi Tian, Chunhua Shen, Hao Chen, and Tong He.
\newblock Fcos: Fully convolutional one-stage object detection.
\newblock In \emph{Proceedings of the IEEE/CVF International Conference on
  Computer Vision (ICCV)}, October 2019.

\bibitem[Uijlings et~al.(2013)Uijlings, Van De~Sande, Gevers, and
  Smeulders]{uijlings2013selective}
Jasper~RR Uijlings, Koen~EA Van De~Sande, Theo Gevers, and Arnold~WM Smeulders.
\newblock Selective search for object recognition.
\newblock \emph{International journal of computer vision}, 2013.

\bibitem[Vaswani et~al.(2017)Vaswani, Shazeer, Parmar, Uszkoreit, Jones, Gomez,
  Kaiser, and Polosukhin]{vaswani2017attention}
Ashish Vaswani, Noam Shazeer, Niki Parmar, Jakob Uszkoreit, Llion Jones,
  Aidan~N Gomez, {\L}ukasz Kaiser, and Illia Polosukhin.
\newblock Attention is all you need.
\newblock \emph{Advances in neural information processing systems}, 2017.

\bibitem[Wang \& Isola(2020)Wang and Isola]{wang2020understanding}
Tongzhou Wang and Phillip Isola.
\newblock Understanding contrastive representation learning through alignment
  and uniformity on the hypersphere.
\newblock In \emph{International Conference on Machine Learning}. PMLR, 2020.

\bibitem[Wang et~al.(2021{\natexlab{a}})Wang, Zhang, Shen, Kong, and
  Li]{wang2021dense}
Xinlong Wang, Rufeng Zhang, Chunhua Shen, Tao Kong, and Lei Li.
\newblock Dense contrastive learning for self-supervised visual pre-training.
\newblock In \emph{Proceedings of the IEEE/CVF Conference on Computer Vision
  and Pattern Recognition}, 2021{\natexlab{a}}.

\bibitem[Wang et~al.(2021{\natexlab{b}})Wang, Zhang, Yang, and
  Sun]{wang2021anchor}
Yingming Wang, Xiangyu Zhang, Tong Yang, and Jian Sun.
\newblock Anchor {DETR}: Query design for transformer-based object detection.
\newblock \emph{arXiv preprint arXiv:2109.07107}, 2021{\natexlab{b}}.

\bibitem[Wei et~al.(2021)Wei, Gao, Wu, Hu, and Lin]{wei2021aligning}
Fangyun Wei, Yue Gao, Zhirong Wu, Han Hu, and Stephen Lin.
\newblock Aligning pretraining for detection via object-level contrastive
  learning.
\newblock \emph{Advances in Neural Information Processing Systems}, 2021.

\bibitem[Wu et~al.(2018)Wu, Xiong, Yu, and Lin]{wu2018unsupervised}
Zhirong Wu, Yuanjun Xiong, Stella~X Yu, and Dahua Lin.
\newblock Unsupervised feature learning via non-parametric instance
  discrimination.
\newblock In \emph{Proceedings of the IEEE conference on computer vision and
  pattern recognition}, 2018.

\bibitem[Xiao et~al.(2021)Xiao, Reed, Wang, Keutzer, and
  Darrell]{xiao2021region}
Tete Xiao, Colorado~J Reed, Xiaolong Wang, Kurt Keutzer, and Trevor Darrell.
\newblock Region similarity representation learning.
\newblock In \emph{Proceedings of the IEEE/CVF International Conference on
  Computer Vision}, 2021.

\bibitem[Xie et~al.(2021)Xie, Lin, Zhang, Cao, Lin, and Hu]{xie2021propagate}
Zhenda Xie, Yutong Lin, Zheng Zhang, Yue Cao, Stephen Lin, and Han Hu.
\newblock Propagate yourself: Exploring pixel-level consistency for
  unsupervised visual representation learning.
\newblock In \emph{Proceedings of the IEEE/CVF Conference on Computer Vision
  and Pattern Recognition}, 2021.

\bibitem[Yang et~al.(2021)Yang, Wu, Zhou, and Lin]{yang2021instance}
Ceyuan Yang, Zhirong Wu, Bolei Zhou, and Stephen Lin.
\newblock Instance localization for self-supervised detection pretraining.
\newblock In \emph{Proceedings of the IEEE/CVF Conference on Computer Vision
  and Pattern Recognition}, 2021.

\bibitem[Yang et~al.(2022)Yang, Huang, and Wang]{yang2022querydet}
Chenhongyi Yang, Zehao Huang, and Naiyan Wang.
\newblock Querydet: Cascaded sparse query for accelerating high-resolution
  small object detection.
\newblock In \emph{Proceedings of the IEEE/CVF Conference on Computer Vision
  and Pattern Recognition}, 2022.

\bibitem[Zheng et~al.(2021)Zheng, You, Wang, Qian, Zhang, Wang, and
  Xu]{zheng2021ressl}
Mingkai Zheng, Shan You, Fei Wang, Chen Qian, Changshui Zhang, Xiaogang Wang,
  and Chang Xu.
\newblock Ressl: Relational self-supervised learning with weak augmentation.
\newblock \emph{Advances in Neural Information Processing Systems}, 2021.

\bibitem[Zhu et~al.(2021)Zhu, Su, Lu, Li, Wang, and Dai]{zhu2020deformable}
Xizhou Zhu, Weijie Su, Lewei Lu, Bin Li, Xiaogang Wang, and Jifeng Dai.
\newblock Deformable {DETR}: Deformable transformers for end-to-end object
  detection.
\newblock In \emph{International Conference on Learning Representations}, 2021.

\end{thebibliography}
\bibliographystyle{iclr2023_conference}

\newpage 

\appendix

\section{Datasets and evaluation details}\label{ax:protocols}

We use different datasets throughout this work that we present below.
\begin{itemize}
    \item For pretraining purposes, we use the standard \emph{ImageNet ILSVRC 2012 (IN)} \citep{russakovsky2015imagenet} dataset, which contains 1.2M training images separated in 1000 class categories.
    \item We also use the \emph{MS-COCO (COCO)} \citep{lin2014microsoft} dataset for pretraining, but mainly for finetuning and evaluation purposes. This dataset contains 80 classes of objects and about 118k training images \emph{(train2017 subset)}. Performance is evaluated on the \emph{val2017} subset.
    \item For finetuning, we also use the \emph{Pascal VOC 2007 and 2012 (VOC 07-12)} \citep{everingham2010pascal} datasets. This dataset contains 20 classes of objects, and we use the combination of the \emph{trainval} subsets from both VOC2007 and VOC2012 for training, corresponding to about 16k training images in total. Performance is evaluated on the \emph{test} subset from VOC2007.
    \item For a more complicated dataset, we use the \emph{Few-Shot Object Detection} dataset \citep{fan2020few}. Since the dataset is designed as \emph{open-set}, \ie with different classes between training and testing, we \rev{separately} use the \rev{\emph{train} and} \emph{test} sets for benchmarking. We separate the test set into a training and testing subset, by randomly taking 80\% of images for training and the 20\% remaining for testing, \rev{and do the same for the train set}. We make sure that all classes appears at least once in both training and testing subsets. The images selected for training and testing will be made available for reproducibility. This separation leads to about 11k training images and 3k testing images \rev{for the \emph{test set}, and about 42k training images and 10k testing images for the \emph{train} set}.
\end{itemize}

\section{Augmentations used}\label{ax:augm}

We detail in \cref{tab:augmentations} the distributions of augmentations used to create the weak view and the strong view. The \emph{Weak Augmentations} follow standard supervised training for transformer-based detectors~\citep{carion2020end,zhu2020deformable}. The \emph{Strong Augmentations} follow typical contrastive learning augmentations \citep{chen2020simple, bar2022detreg}.

\begin{table}[ht]
        \begin{center}
        \caption{The different sets of augmentations used for each branch (\emph{weak} or \emph{strong}). \emph{Probability} indicates the probability of applying the corresponding augmentation.}
        \label{tab:augmentations}
        \end{center}
    \centering
    \begin{tabular}{@{}lcc@{}}
    \toprule
    \midrule
    \multicolumn{3}{c}{\textbf{Weak Augmentations ($\mathcal{T}^1$)}} \\
    \midrule
    Augmentations & Probability & Parameters \\
    \midrule
    Horizontal Flip & 0.5 & -- \\
    \midrule
    \multirow{2}{*}{Resize} & \multirow{2}{*}{0.5} & \emph{Mid-scale:} short edge = range(320,481,16) \\
    & & \emph{Large-scale:} short edge = range(480,801,32) \\
    \midrule
    Resize & \multirow{4}{*}{0.5} & short edge $\in \{400,500,600\}$ \\
        \cmidrule(r){1-1}
        \cmidrule(l){3-3}
    Random Size Crop & & min size = 384 ; max size = 600 \\
        \cmidrule(r){1-1}
        \cmidrule(l){3-3}
    \multirow{2}{*}{Resize} & & \emph{Mid-scale:} short edge = range(320,481,16) \\
    & & \emph{Large-scale:} short edge = range(480,801,32) \\
    \bottomrule
    \toprule
    \multicolumn{3}{c}{\textbf{Strong Augmentations ($\mathcal{T}^2$)}} \\
    \midrule
    Color Jitter & 0.8 & (brightness, contrast, saturation, hue) = (0.4, 0.4, 0.4, 0.1) \\
    \midrule
    GrayScale & 0.2 & -- \\
    \midrule
    Gaussian Blur & 0.5 & (sigma x, sigma y) = (0.1, 2.0) \\
    \bottomrule
    \end{tabular}
\end{table}

\section{\rev{Pretraining cost}}\label{ax:cost}

\rev{
We compare the cost of pretraining \emph{in terms of memory and hardware used} to SoCo \citep{wei2021aligning} in \cref{tab:pt_cost} since it is the closest in terms of pretraining pipeline. The information are derived from their paper and official github repository.}

\begin{table}[ht]
    \begin{center}
        \caption{Comparison of pretraining cost between overall pretraining methods. We compare the number of pretraining epochs on IN, the total batch size, the total number of iterations, the total training time (in hours), the number of iterations per seconds (It. / sec.), and the hardware used (number and type of GPUs). We can see that our pretraining is globally less costly than SoCo. }
        \label{tab:pt_cost}
    \end{center}
    \centering
    \begin{tabular}{@{}lcccccc@{}}
    \toprule
    Method & IN epochs & Batch Size & Iterations & Time & It. / sec. & Hardware \\
    \midrule
    SoCo & 400 & 2048 & 240k & 140h & 0.5 & 16 V100 32G \\
    ProSeCo (Ours) & 10 & 64 & 187k & 40h & 1.4 & 8 A100 40G \\
    \bottomrule
    \end{tabular}
\end{table}


\rev{Even though A100 GPUs are faster than V100 GPUs, we are \emph{training much faster} which is partly explained by the fact that they learn the backbone along with the detection heads during pretraining, leading to more parameters to learn and more computations.
Furthermore, our ProSeCo requires a smaller batch size leading to less memory and thus less GPUs needed.}

\section{Using another contrastive loss}\label{ax:nce}

The popular InfoNCE loss~\citep{oord2018representation} used for contrastive learning is a similarity based function scaled by the temperature $\tau$ that maximizes agreement between the positive pair of instances and push negatives away. 
However, it suffers from the \emph{class collision problem} \citep{cai2020all, denize2023similarity}, where semantically close instances can be used as negatives in the loss computation, which damages the quality of the representation learned. Recent work \citep{zheng2021ressl, denize2023similarity} have tackled this problem by introducing the relational aspect between instances. 
In our experimental study, we also considered the InfoNCE loss for ProSeCo. We formulate the loss as follows: 

\begin{equation}
    \Lcal_\text{InfoNCE}(\rvz, \hat{\rvz}, \hat{\sigma}^\text{prop}) = - \frac{1}{N_b N} \sum_{i=1}^{N_b} \sum_{j=1}^N \log \left( \frac{\exp(\rvz_{(i,j)} \cdot \hat{\rvz}_{(i, \hat{\sigma}_i^\text{prop}(j))} / \tau )}{ \sum_{k=1}^{N_b} \sum_{l=1}^{N} \exp(\rvz_{(i,j)} \cdot \hat{\rvz}_{(k,l)} / \tau) } \right).
\end{equation} 

This formulation leads to an effective batch size of $N_b \cdot N$ for the contrastive loss.
Similarly to our \emph{LocSCE}, the localization of the objects can be introduced in the InfoNCE loss function to obtain the \emph{LocNCE} objective as follows: 
\begin{equation}
    \Lcal_\text{LocNCE}(\rvy, \hat{\rvz}, \hat{\sigma}^\text{prop}) = - \frac{1}{N_b N} \sum_{i=1}^{N_b} \sum_{j=1}^N \sum_{m=1}^N \1_{IoU_i(j,m) \geq \delta} \log \left( \frac{ \exp(\rvz_{(i,j)} \cdot \hat{\rvz}_{(i, \hat{\sigma}_i^\text{prop}(m))} / \tau )}{ \sum_{k=1}^{N_b} \sum_{l=1}^{N} \exp(\rvz_{(i,j)} \cdot \hat{\rvz}_{(k,l)} / \tau) } \right).
\end{equation} 

In \cref{tab:abl_loss}, we compare the two different contrastive objectives for pretraining. We can see that using the InfoNCE loss leads to slightly better results (+0.3 p.p.). However, when using the localization information, SCE benefits much more than InfoNCE (+1.7 p.p. compared to +0.6 p.p.). This might be that the selection of positives from the localization information helps to introduce easy positive examples, and thanks to this, the relational aspect of SCE can focus on the more difficult positives. In the end, LocSCE achieves stronger results than LocNCE (+0.8 p.p.).

\begin{table}[ht]
    \begin{center}
        \caption{Performance (mAP in \%) comparison on Mini-COCO 5\% of the different contrastive loss and the effect of the localization information on each of them.}
        \label{tab:abl_loss}
    \end{center}
    \centering
    \begin{tabular}{@{}lcc@{}}
    \toprule
    Loss & $\delta$ & mAP \\
    \midrule
        InfoNCE & 1.0 & 26.4 \\
        \textit{LocNCE (Ours)} & 0.5 & \textbf{27.0} \\
    \midrule
        SCE & 1.0 & 26.1 \\
        \textit{LocSCE (Ours)} & 0.2 & 27.0 \\
        \textit{LocSCE (Ours)} & 0.7 & 27.1 \\
        \textit{LocSCE (Ours)} & 0.5 & \textbf{27.8} \\
    \bottomrule
    \end{tabular}
\end{table}

\section{Full evaluation metrics}\label{ax:full_eval}

\cref{tab:mini_voc,tab:fsod} provide the results with full evaluation metrics (mAP, $\text{AP}_{50}$ and $\text{AP}_{75}$ in \%) on PASCAL VOC, Mini-VOC, FSOD-test \rev{and FSOD-train} benchmarks.

\begin{table}[ht]
        \begin{center}
        \caption{Performance comparison after finetuning on PASCAL VOC and in the novel Mini-VOC setting. On Mini-VOC, we use different percentage of training data (with the corresponding number of images reported) for finetuning.}
        \label{tab:mini_voc}
    \end{center}
    \resizebox{0.99\linewidth}{!}{%
    \begin{tabular}{@{}lccccccccc@{}}
    \toprule
    \multirow{3}{*}{Method} & \multicolumn{3}{c}{PASCAL VOC} & \multicolumn{6}{c}{Mini-VOC} \\
    \cmidrule(lr){5-10}
    & \multicolumn{3}{c}{100\% (16k)} & \multicolumn{3}{c}{5\% (0.8k)} & \multicolumn{3}{c}{10\% (1.6k)} \\
    \cmidrule(lr){2-4}
    \cmidrule(lr){5-7}
    \cmidrule(lr){8-10}
    & mAP & $\text{AP}_{50}$ & $\text{AP}_{75}$ & mAP & $\text{AP}_{50}$ & $\text{AP}_{75}$ & mAP & $\text{AP}_{50}$ & $\text{AP}_{75}$ \\
    \midrule
        
        Supervised & 59.5 & 82.6 & 65.6 & 33.9 & 56.9 & 35.0 & 40.8 & 63.7 & 43.1 \\
        DETReg \citep{bar2022detreg} & 63.5 & 83.3 & 70.3 & 43.1 & 63.4 & 46.1 & 48.2 & 68.6 & 51.9 \\
        \textit{ProSeCo (Ours)} & \textbf{65.1} & \textbf{84.7} & \textbf{73.0} & \textbf{46.1} & \textbf{66.1} & \textbf{50.2} & \textbf{51.3} & \textbf{72.7} & \textbf{56.1} \\
    \bottomrule
    \end{tabular}%
    }
\end{table}



\begin{table}[ht]
        \begin{center}
        \caption{Performance comparison on FSOD-test \rev{and FSOD-train}}
        \label{tab:fsod}
    \end{center}
    \centering
    \begin{tabular}{@{}lcccccc@{}}
    \toprule
    \multirow{3}{*}{Method} & \multicolumn{3}{c}{FSOD-test (11k)} & \multicolumn{3}{c}{\rev{FSOD-train (42k)}} \\
    \cmidrule(lr){2-4}
    \cmidrule(lr){5-7}
    & mAP & $\text{AP}_{50}$ & $\text{AP}_{75}$ & \rev{mAP} & \rev{$\text{AP}_{50}$} & \rev{$\text{AP}_{75}$} \\
    \midrule
        Supervised & 39.3 & 57.7 & 42.6 & \rev{42.6} & \rev{58.1} & \rev{46.5} \\
        DETReg \citep{bar2022detreg} & 43.2 & 59.6 & 47.8 & \rev{43.3} & \rev{57.9} & \rev{47.3} \\
        \textit{ProSeCo (Ours)} & \textbf{46.6} & \textbf{64.5} & \textbf{50.9} & \rev{\textbf{47.2}} & \rev{\textbf{62.4}} & \rev{\textbf{51.7}} \\
    \bottomrule
    \end{tabular}
\end{table}

\section{Increasing the number of queries}\label{ax:queries}

In \cref{tab:abl_queries}, we provide an ablation on the number of object proposals (queries) $N$ in \ddetr, when pretraining with ProSeCo and finetuning afterward. 
A higher $N$ leads to more parameters in the model and longer computing time, \rev{but we can see that the results of \ddetr are relatively stable w.r.t. to the number of queries}. On the other hand, ProSeCo benefits from increasing the number of queries, since it means a higher effective batch size during contrastive learning. However, the default value of $N=300$ leads to the best results, both with and without pretraining.

\begin{table}[ht]
        \begin{center}
        \caption{Performance (mAP in \%) comparison on Mini-COCO 5\% when changing the number of object proposals in \ddetr.}
        \label{tab:abl_queries}
    \end{center}
    \centering
    \begin{tabular}{@{}lcc@{}}
    \toprule
    Method & $N$ & Performance \\
    \midrule
        \multirow{4}{*}{Supervised} & 100 & 23.1 \\
        & 200 & 23.0 \\
        & 300 & \textbf{23.6} \\
         & 500 & 23.3 \\
     \midrule
        \multirow{4}{*}{\textit{ProSeCo (Ours)}} & 100 & 25.7 \\
        & 200 & 26.5 \\
        & 300 & \textbf{27.8} \\
         & 500 & 27.2 \\
    \bottomrule
    \end{tabular}
\end{table}

\section{\rev{Finetuning with a lot of data}}\label{ax:ft_full_coco}

\rev{In \cref{tab:full_coco}, we present results when finetuning on the full COCO dataset under the $1\times$ training schedule \citep{wei2021aligning, li2022dn}, \ie 12 training epochs and decaying the learning rate in the last epoch. The improvements in the large-scale annotated data regime are limited, which can be observed also in previous work \citep{dai2021up, bar2022detreg}. As we can see, our ProSeCo reaches similar results than DETReg \citep{bar2022detreg}. We believe that this limitation comes from the pretrained backbone that stays fixed during pretraining, and from the extensive supervision during fine-tuning. However, as we can see in both \cref{tab:fsod,tab:voc_fsod}, we outperform DETReg on our \emph{FSOD-train} benchmark, which represents a setting with mid-scale annotated data (42k training images).}

\begin{table}[ht]
        \begin{center}
        \caption{\rev{Performance (mAP, $\text{AP}_{50}$ and $\text{AP}_{75}$ in \%) comparison on the full COCO dataset (118k training images) with the $1\times$ training schedule.}}
        \label{tab:full_coco}
    \end{center}
    \centering
    \begin{tabular}{@{}lccc@{}}
    \toprule
    \multirow{2}{*}{Method} & \multicolumn{3}{c}{COCO (118k)}\\
    \cmidrule(lr){2-4}
    & mAP & $\text{AP}_{50}$ & $\text{AP}_{75}$ \\
    \midrule
        Supervised & 37.4 & 55.5 & 40.5 \\
        DETReg \citep{bar2022detreg} & 38.9 & 56.6 & 42.3 \\
        \textit{ProSeCo (Ours)} & 38.9 & 56.2 & 42.4 \\
    \bottomrule
    \end{tabular}
\end{table}

\section{\rev{Performance comparison of Detectors in the few data regime}}\label{ax:perf_comp_fsl}

\rev{From the results presented in \cref{tab:comp_ddetr_frcnn}, we can see that \ddetr, a recent state-of-the-art detection model based on transformers, achieves consistently better performance than the most popular two-stage method in when learning with limited labels. These differences in performance are all the more impressive since the two methods have a similar number of parameters. These results motivated our choice of transformer-based architectures for our pretraining method.}

\begin{table}[ht]
    \begin{center}
        \caption{Performance (mAP in \%) comparison between Faster-RCNN (FRCNN)~\citep{ren2015faster} with Feature Pyramid Network (FPN)~\citep{lin2017feature}, a two-stage detector commonly used, FCOS~\citep{tian2019fcos}, a more recent one-stage method, and Deformable DETR (\ddetr)~\citep{zhu2020deformable}, a state-of-the-art transformer-based object detector, with the same ResNet-50 backbone model. \\
        $^\dagger$ Results from \citet{Liu_2022_CVPR}. $^\ddagger$ Results from \citet{Bouniot_2023_WACV}.
        }
        \label{tab:comp_ddetr_frcnn}
    \end{center}
    \centering
        \begin{tabular}{@{}lcccc@{}}
        \toprule
        \multirow{2}{*}{Method} & \multicolumn{4}{c}{Mini-COCO} \\
        \cmidrule(lr){2-5}
         & 0.5\% (590) & 1\% (1.2k) & 5\% (5.9k) & 10\% (11.8k) \\
        \midrule
            FCOS$^\dagger$ & $5.42 \pm 0.01$ & $8.43 \pm 0.03$ & $17.01 \pm 0.01$ & $20.98 \pm 0.01$ \\
            FRCNN + FPN$^\dagger$ & $6.83 \pm 0.15$ & $9.05 \pm 0.16$ & $18.47 \pm 0.22$ & $23.86 \pm 0.81$ \\
            Def.~DETR$^\ddagger$ & $\mathbf{8.95} \pm \mathbf{0.51}$ & $\textbf{12.96} \pm \textbf{0.08}$ & $\textbf{23.59} \pm \textbf{0.21}$ & $\textbf{28.55} \pm \textbf{0.08}$ \\
        \bottomrule
        \end{tabular}
\end{table}




\end{document}